\documentclass[preprint, 12pt, authoryear]{elsarticle}
\usepackage{amsmath,amssymb,amsfonts}
\usepackage{algorithmic}
\usepackage{graphicx}
\usepackage{textcomp}
\usepackage{amsmath}
\usepackage{hyphenat}
\usepackage[cjk]{kotex}
\usepackage{multirow}
\usepackage{booktabs}
\usepackage{tabularx}
\usepackage{colortbl}
\usepackage{graphicx}
\usepackage{bbding}
\usepackage{pifont}
\usepackage{wasysym}
\usepackage{amssymb}
\usepackage{pdfrender}
\usepackage{wrapfig}
\usepackage{subcaption}
\usepackage{multirow}
\usepackage{makecell}
\usepackage[pagebackref,breaklinks,colorlinks]{hyperref}
\usepackage{tabularx}
\usepackage{graphicx,scalerel}

\newcolumntype{C}{>{\centering\arraybackslash}X}
\usepackage[capitalize]{cleveref}
\Crefname{section}{Section}{Sections}
\crefname{section}{Sec.}{Secs.}
\Crefname{table}{Table}{Tables}
\crefname{table}{Tab.}{Tabs.}
\Crefname{equation}{Equation.}{Equations.}
\crefname{equation}{Eq.}{Eqs.}

\newcommand\clearrow{\global\let\rowmac\relax}

\def\SP#1{\textsuperscript{#1}}

\newcommand{\ie}{\textit{i}.\textit{e}., }

\usepackage[T1]{fontenc}
\usepackage{graphicx, kotex, mathtools, amssymb, booktabs}
\graphicspath{{./img/}}
\hypersetup{colorlinks = true, linkcolor = blue, anchorcolor = red, citecolor = blue, filecolor = red, urlcolor = red, pdfauthor=author}

\journal{Computer Methods and Programs in Biomedicine}
\begin{document}
\begin{frontmatter}
\title{Fine-Grained Self-Supervised Learning with Jigsaw Puzzles for Medical Image Classification}
\author[ajou1]{Wongi Park}
\author[ajou1,ajou2]{Jongbin Ryu\corref{corres}}
\cortext[corres]{Corresponding author. Jongbin Ryu (e-mail: jongbinryu@ajou.ac.kr)}
\address[ajou1]{Department of Software, Ajou University}
\address[ajou2]{Department of Computer Engineering, Ajou University}

\begin{abstract}
Classifying fine-grained lesions is challenging due to minor and subtle differences in medical images. 
This is because learning features of fine-grained lesions with highly minor differences is very difficult in training deep neural networks. 
Therefore, in this paper, we introduce \textbf{F}ine-\textbf{G}rained \textbf{S}elf-\textbf{S}upervised \textbf{L}earning(\textbf{FG-SSL}) method for classifying subtle lesions in medical images.
The proposed method progressively learns the model through hierarchical block such that the cross-correlation between the fine-grained Jigsaw puzzle and regularized original images is close to the identity matrix. We also apply hierarchical block for progressive fine-grained learning, which extracts different information in each step, to supervised learning for discovering subtle differences. Our method does not require an asymmetric model, nor does a negative sampling strategy, and is not sensitive to batch size. 
We evaluate the proposed fine-grained self-supervised learning method on comprehensive experiments using various medical image recognition datasets.
In our experiments, the proposed method performs favorably compared to existing state-of-the-art approaches on the widely-used ISIC2018, APTOS2019, and ISIC2017 datasets.

\begin{keyword}
Self-Supervised learning, Fine-grained medical image recognition, Jigsaw puzzle, Progressive learning
\end{keyword}



\end{abstract}
\end{frontmatter}
\section{Introduction}
Self-supervised learning has a lot of potential due to its general representation of learning without supervision.
Recently, numerous studies \cite{kaczmarzyk2023champkit, zhao2023bidirectional, wargnier2023weakly, jiao2023gmrlnet, yang2023hierarchical, yu2023bayesian, chen2023fit} 
show that a pre-trained model with self-supervised learning can successfully transfer the general knowledge to the medical image analysis for computer-aided diagnosis (CAD). 
Specifically, approaches of learning two different models asymmetrically\cite{liu2022act} and metric learning with triplet sampling strategy\cite{yang2022proco} have been actively studied in the medical image recognition. 
These approaches focus on unsupervised self-labeling of unchanging information by sampling or contrastive learning strategy. 
Therefore, creating discriminant and consistency self-labeling has been the most critical issue.
In this paper, we go a step further by introducing fine-grained self-supervised learning for medical image recognition.
The proposed FG-SSL tackles extracting fine-grained information in the process of self-supervised learning so that we can further learn the subtle pathological feature in medical images. 
Our FG-SSL compares the distorted image to shuffled Jigsaw puzzles with different granularity levels, in which a model with shared weights gradually learns the self-supervision of Jigsaw puzzles. This learning process allows the network to learn information from small to relatively big areas simultaneously from the progressive Jigsaw puzzle. In addition, when building shuffled Jigsaw puzzles, we apply an effective and strong distortion\cite{kim2018learning}, forcing the network to learn diversified features.
The followings are the primary contributions of this paper.

\setlength{\topsep}{0pt}
\setlength{\itemsep}{0pt}
\setlength{\partopsep}{0pt}
%
\begin{itemize}
\item 
We introduce a fine-grained self-supervised learning method for recognizing subtle differences in medical images by minimizing the cross-correlation between original and shuffled images of Jigsaw puzzles.
\item 
We enhance our self-supervised learning with the extensive data augmentation of original images. We regularize the original images to learn the diversified representation in our self-supervised learning. 
\item
We present a straightforward learning model that does not necessitate an asymmetric model, a negative sampling strategy, and a large batch size. This ensures that the proposed method is applicable to real-world medical image classification tasks.
\item 
We demonstrate that the proposed fine-grained self-supervised method shows state-of-the-art performance on the ISIC2018, APTOS2019, and ISIC2017 datasets. Furthermore, each component of the proposed method is evaluated by an extensive ablation study.
\end{itemize}
%

\section{Method}

\begin{figure*}[ht]
\centering
\includegraphics[width=0.985\textwidth]{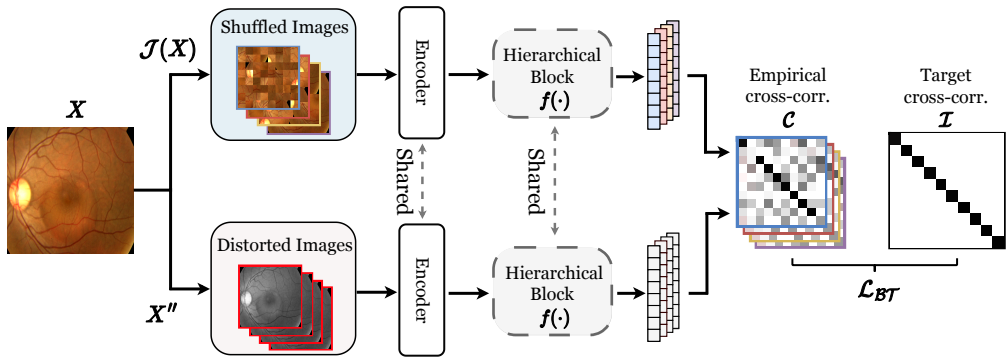}
\caption{ Overall workflow of the proposed FG-SSL method. We generate Jigsaw puzzles and distorted images from an original training sample to learn fine-grained self-supervision. By comparing the features obtained from fine-grained Jigsaw puzzles and distorted images, we minimize the empirical cross-correlation with the identity matrix for progressively training our networks.
}
\label{fig:FG_SSL}
\end{figure*}

In this section, we present a method dubbed \textbf{F}ine-\textbf{G}rained \textbf{S}elf-\textbf{S}upervised \textbf{L}earning (\textbf{FG-SSL}) for medical image recognition. The overall workflow of the proposed method is shown in Fig.~\ref{fig:FG_SSL}. In Sec.~\ref{Self-supervision}., we first discuss the basis of this work regarding the single-level Jigsaw puzzle problem, and introduce our multi-level fine-grained self-supervised learning with a progressive Jigsaw puzzle. In Sec.~\ref{Regularizing}., a regularization approach is also provided to improve the proposed fine-grained self-supervised method, and lastly, the step of fine-tuning employing the self-supervised pretrained model for classifying lesion classes of the medical data.

\subsection{Self-supervision with Progressive Jigsaw puzzle}
\label{Self-supervision}
\textbf{Single-level Jigsaw puzzle}. The Jigsaw puzzle problem has been widely used in various self-supervised learning methods\cite{liu2021simtriplet, zhuang2019self, li2020self}.
Since it can learn the unique form of the original image without labeling, self-supervised learning helps supervised learning as a pretext.
Before applying the target task with supervised learning, a network is pretrained with the self-supervision of the single-level Jigsaw puzzle.
The original image is compared to the Jigsaw puzzle that is shuffled in the single-level grid, as shown in Fig.~\ref{fig:FG_SSL}. 
This single-level Jigsaw puzzle for self-supervised learning is formulated as: \begin{equation}
\label{eq:formula_jigsaw}
p(S|L_{1}, ...., L_{N}) = p(S|F_{1}, F_{2}, F_{1}, ...., F_{N}) = \prod_{i=1}^{N}p(L_{i}|F_{i}),
\end{equation} 
where $F_{i}$ denotes features extracted from a Jigsaw patch $L_{i}$, $S$ represents the shuffling order of Jigsaw patches, and $N$ stands for the number of patches.\cite{noroozi2016unsupervised}
Although this single-level Jigsaw puzzle shows competitive performance, we argue that it is limited to learning fine-grained information about medical images, especially lesion classes. 

\begin{figure*}[!t]
\centering
\includegraphics[width=0.995\textwidth]{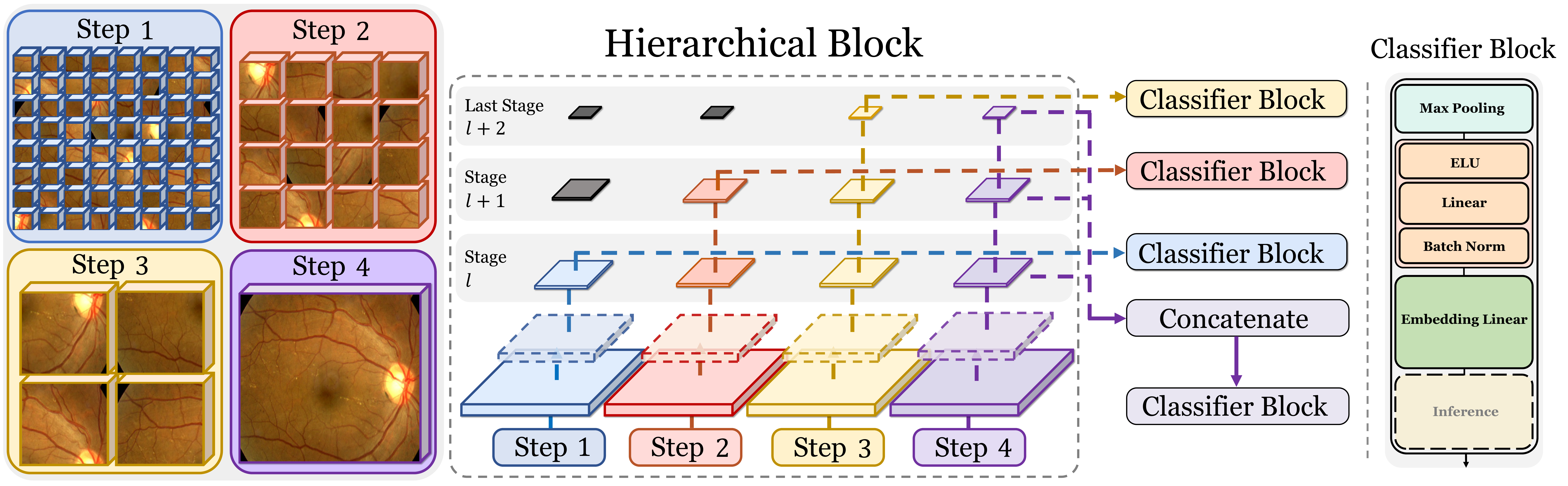}
\caption{
The hierarchical structure of our self-supervised learning method. At each step of the hierarchical blocks, we progressively train the networks using shuffled Jigsaw patches with different granularity levels.}
\label{fig:hierarchical_block}
\end{figure*}

\noindent \textbf{Progressive multi-level Jigsaw puzzle.}
Therefore, we introduce the multi-level Jigsaw puzzle to a progressive training strategy. 
We assume that at the stage level of a network, distinct self-supervisions should be learned hierarchically in order to understand the generic form of an object; early-stage layers should encode detailed low-level information, while late-stage layers should consider high-level abstracted features. 
For this purpose, we use the recently proposed PMG~\cite{du2020fine} for our multi-level self-supervised learning method. The PMG trains a network with different granularity levels of the Jigsaw puzzle.
We apply the progressive Jigsaw puzzles of the PMG model in our self-supervised learning by shuffling the puzzles with different levels, as shown in Fig.~\ref{fig:hierarchical_block}.
In the early stages of a network, the distorted images are compared to a Jigsaw puzzle shuffled in a grid size of ${N \times N}$, while subsequent stages employ more abstracted grid sizes such as $(N/2 \times N/2), (N/4 \times N/4)$, and $(N/8 \times N/8)$.
To learn the self-supervision from the multiple Jigsaw puzzles, we extract PMG features from each stage of a network as:

\begin{equation}
\label{eq:ext_jig_feat}
\hat{\mathcal{X}}^{l} = \psi[\mathcal{X}_{1}^{l}, \mathcal{X}_{2}^{l}, ..., \mathcal{X}_{N}^{l}], \quad \text{where} \quad \mathcal{X}_{i}^{l} = H_{conv}^{l}(F_{i}^{l}),
\end{equation}
where $l$ denotes the index of a stage, $H_{conv}$ represents the convolution layer to extract the patch feature $\mathcal{X}_{i}$, and
$\psi[\cdot]$ is the concatenation function of multiple feature maps which generate the stage feature $\hat{\mathcal{X}}$.  We then train a network with self-supervised learning as 
\begin{equation} 
\label{eq:general_block}
\mathcal{L}_{CE} = \sum_l{y\times\log{y_{i}^{l}}},  \quad \text{where} \quad y^{l} = H^{l}(\hat{\mathcal{X}}^{l}), 
\end{equation}


$\mathcal{L}_{CE}$ is the cross-entropy loss, $y$ and $y^{l}$ represent the ground truth and our empirical label of the Jigsaw puzzle ordering from each stage, $H(\cdot)$ denotes the classifier layer. We compute the final loss with the cross-entropy on each stage of a network progressively: $\mathcal{L}_{CE} = \sum_{l}{\mathcal{L}^{l}_{CE}(y^{l}, y)}$.


Using this progressive learning process, our network can be pre-trained with finely shuffled Jigsaw puzzles.
This fine-grained self-supervised learning has the advantage of learning various fine-grained hierarchical features in multi-stage Jigsaw puzzles.
\subsection{Augmentation for Regularizing Self-supervision}
\label{Regularizing}
Whereas the previous subsection introduced the SSL method of a target Jigsaw puzzle, this subsection provides the extensive fine-grained augmentation method of an original image directly for enhancing our self-supervision ability. 
We use the BarlowTwins\cite{zbontar2021barlow} method for fine-grained augmentation of original images, which uses the $negative\text{-}sample\text{-}free$ approach. BarlowTwins learn self-supervision by using random augmentation on the original image to minimize cross-correlation with the shuffled jigsaw images. 
Therefore, it computes the cross-correlation $C$ between the features of augmented original image $\hat{\mathcal{X}}$ with the Jigsaw puzzle $\mathcal{J}(\mathcal{X})$ as:
\begin{equation}
\label{eq:cross_correlation}
C_{ij} = \frac{\mathbb{E}\big[[f(\hat{\mathcal{X}_{i}})][f(\mathcal{J}(\mathcal{X}_{j}))]\big]}{\sqrt{\mathbb{E}[[f(\hat{\mathcal{X}_{i}})]^2]}\sqrt{\mathbb{E}[[f(\mathcal{J}(\mathcal{X}_{j}))]^2]}},
\end{equation}
where $f(\cdot)$ represents the feature extractor, $\mathbb{E}$ denotes the expectation function, and $i, j$ stand for the indices of training samples.
Based on this definition of cross-correlation $C$, we define loss function of BarlowTwins for comparing both augmented original and Jigsaw puzzles as 
\begin{equation}
\label{eq:Barlow_twins}
\mathcal{L_{BT}} = \sum\limits_{i}^{}{(1 - \mathcal{C}^{aug, Jigsaw}_{i, i})} \quad  + \lambda \sum\limits_{i}^{}{\sum\limits_{j \neq i}^{}{(\mathcal{C}^{Jigsaw}_{ij}})^2}, \
\end{equation}
where the left term considers the invariance between the augmented original image and its Jigsaw puzzle, and the right term takes into account the redundancy reduction between the augmented original image and with Jigsaw puzzle from a different image. The positive constant $\lambda$, which is a user-input hyperparameter, is used to balance the importance of the first and second terms in the loss function.
The invariance term, in particular, requires cross-correlation of the same images to resemble the identity matrix, resulting in Jigsaw puzzle invariance for fine-grained augmentations. Furthermore, by decreasing the cross-correlation between different images, the redundancy reduction term serves to push them away from the feature space.
We then eventually acquire the pre-trained model by using this loss function of Eq.~\ref{eq:Barlow_twins} under our fine-grained self-supervision.
We dub this fine-grained self-supervised learning in the remaining sections as \textbf{FG-SSL}.

\textbf{Fine-tuning.}
We fine-tune our networks from self-supervised learning for medical classification. With this fine-tuning, we also adopt the progressive Jigsaw puzzle of the PMG model.
However, unlike the self-supervised learning as a pretext, we use the cross entropy(CE) with ground truth and estimated lesion labels as :
\begin{equation}
\label{eq:supervised_combined_class}
\mathcal{L}_{CE}(y^{P}, y) = -\sum\limits_{_i = 1}^{m}y_{i}^{P}\times log(y_{i}^{P}),
\end{equation}
where $y$ and $y^P$ are the ground truth and estimated labels, and $P$ denote the stage of our hierarchical architecture.

\section{Experiments}
\subsection{Dataset and Evaluation}
We evaluate the proposed FG-SSL method with ISIC2018, APTOS2019, and ISIC2017 datasets. 
The ISIC2017 and ISIC2018 are available for download throu-gh the skin detection challenge\cite{8363547}\footnote{https://challenge.isic-archive.com/data/}. The APTOS2019\cite{aptos2019-blindness-detection} dataset can also be found in the public \footnote{https://www.kaggle.com/competitions/aptos2019-blindness-detection/data}. 
The ISIC2017 dataset has two classes with 2000 training and 600 test images.
The ISIC2018 dataset contains seven classes and 10015 skin images, and APTOS2019 has five classes and a total of 3662 diabetic retinopathy images.
In order to compare the performance fairly with existing methods\cite{yang2022proco, wu2020skin}, we evaluate the proposed FG-SSL by randomly dividing the ISIC2018 and APTOS2019 datasets into train and test at a ratio of 7:3, in the same way as in the previous methods. Also, we perform experiments of ISIC2017 Task-3A, i.e., melanoma detection.
We evaluate APTOS2019 and ISIC2018 using accuracy and F1-score, as in previous studies, and ISIC2017 using accuracy and area under the curve (AUC) as the comparative metrics.
On the ISIC2018, APTOS2019, and ISIC2017 datasets, we compare existing methods, including state-of-the-art methods, in Table ~\ref{entire_experimental}.

\begin{table*}[t]
\centering
\renewcommand{\arraystretch}{1.25}
\setlength\heavyrulewidth{0.7pt}
\caption{Experimental comparison with the state-of-the-art-methods.}
\label{entire_experimental}
\resizebox{\textwidth}{!}{%
\begin{tabular}{lccccclcc} 
\cmidrule[\heavyrulewidth]{1-5}
\cmidrule[\heavyrulewidth]{7-9}
\multicolumn{1}{l}{\multirow{2}{*}{Methods}} 
& \multicolumn{2}{c}{\;APTOS2019} 
& \multicolumn{2}{c}{\;ISIC2018} 
& \multicolumn{1}{c}{\;\;} 
& \multicolumn{1}{l}{\multirow{2}{*}{Methods}} 
& \multicolumn{2}{c}{ISIC2017} \\ 
\cmidrule(l){2-3}
\cmidrule(l){4-5}
\cmidrule(l){8-9}
& Acc 
& F1 
& Acc 
& F1 
& \multicolumn{1}{c}{} 
& \multicolumn{1}{c}{} 
& Acc 
& \multicolumn{1}{l}{AUC} \\ 
\cmidrule{1-5}
\cmidrule{7-9}
CE & 0.812 & 0.608 & 0.850 & 0.716 &  & GoogLeNet & 0.764 & 0.767 \\
CE+resample & 0.802 & 0.583 & 0.861 & 0.735 &  & ResNet50 & 0.835 & 0.852 \\
Focal Loss \cite{lin2017focal} & 0.815 & 0.629 & 0.849 & 0.728 &  & ResNet101 & 0.839 & 0.861 \\
OHEM & 0.813 & 0.631 & 0.818 & 0.660 &  & DenseNet101 & 0.842 & 0.861 \\
LDAM \cite{cao2019learning}& 0.813 & 0.620 & 0.857 & 0.734 &  & SDL \cite{zhang2019medical}& 0.830 & 0.830 \\
MTL & 0.813 & 0.632 & 0.811 & 0.667 &  & Galdran $\text{et } al.$ \cite{galdran2017data}& 0.480 & 0.765 \\
DANIL \cite{gong2020distractor}& 0.825 & 0.660 & 0.825 & 0.674 &  & Vasconcelos $\text{et } al.$ \cite{vasconcelos2017increasing}& 0.830 & 0.791 \\
CL \cite{marrakchi2021fighting} & 0.825 & 0.652 & 0.865 & 0.739 &  & Yang $\text{et } al.$ \cite{yang2017novel} & 0.830 & 0.830 \\
CL + resample \cite{marrakchi2021fighting}& 0.816 & 0.608 & 0.868 & 0.751 &  & Diaz $\text{et } al.$ \cite{diaz2017incorporating} & 0.868 & 0.879 \\
ProCo \cite{yang2022proco}   & 0.837 & 0.674 & 0.887 & 0.763 &  & ARDT-DenseNet\cite{wu2020skin} & 0.868 & 0.879 \\
\cmidrule{1-5}\cmidrule{7-9}
\textbf{FG-SSL(Ours)} & \textbf{0.858} & \textbf{0.720} & \textbf{0.897} & \textbf{0.816} &  & \textbf{FG-SSL(Ours)} & \textbf{0.913} & \textbf{0.927} \\
\cmidrule[\heavyrulewidth]{1-5}
\cmidrule[\heavyrulewidth]{7-9}
\end{tabular}
}
\end{table*}

\subsection{Implementation Details}
\label{details}
The proposed FG-SSL uses ResNet50\cite{he2016deep} as the backbone in the same way as the existing methods. 
In the FG-SSL method for self-supervised learning, resize the input image $256 \times 256$ and resize the input image to $224 \times 224$ with a center crop.
We apply the data augmentation method following previous studies\cite{yang2022proco, chen2020improved}.
During transfer learning, we resize the input image 256 and resize the input image to $224 \times 224$ with a center crop. 
We set the hyperparameter $\lambda$ to the one proposed by the existing author\cite{zbontar2021barlow}.
We utilize the features extracted from the 3rd to the last 5th stages in a total of 5 stages of the ResNet50 model for our progressive learning. 
We set a batch size of 256 and used SGD optimizer with a momentum of $0.9$ and weight decay of $0.001$. The initial learning rate of 0.002 and continuously reduced it using a cosine annealing strategy.
We set training epochs as 1000 and trained our model using the PyTorch version 1.11.0 with RTX A5000 GPUs.

\begin{table}[h]
    \caption{Factor analysis of the proposed FG-SSL.}
    \label{tbl:jigsaw_compare}
    \centering
    \begin{subtable}[c]{0.62\textwidth}
        \centering
        \setlength{\tabcolsep}{1pt}
        \renewcommand{\arraystretch}{1.01}   
        \begin{tabular}{cccccc}
            \toprule
            \multirow{2}{*}{Method\;} & \multirow{2}{*}{Model} & \multicolumn{2}{c}{Single-level} & \multicolumn{2}{c}{Multi-level} \\
            \cmidrule(l){3-6}
             &  & Acc & F1 & Acc & F1 \\ 
            \midrule
            SimSiam&\multirow{3}{*}{Baseline}   & 0.860 & 0.732 & 0.865 & 0.743 \\
            Rotation &  & 0.847 & 0.733 & 0.856 & 0.747 \\
            FG-SSL(Ours) &  & 0.856 & 0.759 & 0.873 & 0.764 \\ 
            \midrule
             SimSiam& \multirow{3}{*}{PMG} & 0.867 & 0.764 & 0.886 & 0.798 \\ 
             MoCoV2&  & 0.818 & 0.720 & 0.841 & 0.783 \\ 
             Rotation&  & 0.841 & 0.739 & 0.867 & 0.751 \\ 
            \midrule
            FG-SSL(Ours) & PMG & 0.882 & 0.794 & 0.897 & 0.816  \\
            \bottomrule
        \end{tabular}       
        \subcaption{Progressive Jigsaw puzzles.}
        \label{tab:compare_a}
    \end{subtable}
    \hfill
    \begin{subtable}[c]{0.37\textwidth}
        \centering
        \setlength{\tabcolsep}{1.5pt}
        \renewcommand{\arraystretch}{1.141}
        \begin{tabular}{ccccccccc}
            \toprule
            \multicolumn{4}{c}{Granularity}  &&&& \multicolumn{2}{c}{FG-SSL} \\ \cmidrule{1-4} \cmidrule{8-9}
            1\SP{st} & 2\SP{nd} & 3\SP{rd} & 4\SP{th} &&&& Acc & F1 \\ 
            \midrule
            8& 8 & 8 & 8 &&&& 0.864 & 0.763 \\ 
            8& 8 & 4 & 2 &&&& 0.878 & 0.784 \\ 
            8& 4 & 4 & 4 &&&& 0.867 & 0.775 \\ 
            8& 4 & 2 & 2 &&&& 0.885 & 0.809 \\ 
            \midrule
            8 & 4 & 2 & 1 &&&& 0.897 & 0.816 \\
            \midrule
            16 & 8 & 4 & 1 &&&& 0.884 & 0.797 \\ 
            \bottomrule
        \end{tabular}
        \subcaption{Progressive Strategy.}
        \label{tab:compare_b}
    \end{subtable}
    \hfill
    \begin{subtable}[c]{0.5\textwidth}
        \centering
        \setlength{\tabcolsep}{1.0pt}
        \renewcommand{\arraystretch}{0.95}
        \begin{tabular}{ccccccc} 
            \toprule
            \multirow{2}{*}{\;Method\;}& &\multicolumn{3}{c}{Batch size} \\ \cmidrule{2-6}
                                       & 16    & 32    &  64   & 128   & 256 \\ \midrule
            MoCoV2 & 0.785 & 0.793 & 0.812 & 0.834 & 0.841 \\ 
            FG-SSL(Ours) & 0.874 & 0.878 & 0.882 & 0.891 & 0.897  \\
            \bottomrule
        \end{tabular}         
        \subcaption{Batch size.}
        \label{tab:compare_c}
    \end{subtable}
    \hfill
    \begin{subtable}[c]{0.43\textwidth}
        \centering
        \setlength{\tabcolsep}{1pt}
        \renewcommand{\arraystretch}{1.17}
        \begin{tabular}{ccccc}
            \toprule    
                Jigsaw & PMG & Barlow & Acc & F1  \\ \hline
                \checkmark &  &  & 0.862 & 0.742  \\ 
                \checkmark & \checkmark &  & 0.883 & 0.779  \\
                \textbf{\checkmark} & \textbf{\checkmark} & \textbf{\checkmark} & 0.897 & 0.816 \\ 
             \bottomrule
        \end{tabular}
        \subcaption{SSL component.}
        \label{tab:compare_d}
    \end{subtable}
    \hfill
    \begin{subtable}[c]{\textwidth}
        \centering
        \setlength{\tabcolsep}{1pt}
        \renewcommand{\arraystretch}{0.8}
        \begin{tabular}{lccccc}
            \toprule
                 Dimension\;\;\; & \;\;  \;\;16  \;\;\;\;\;\;& \;\;\;\; \;\;32  \;\;\;\;\;\; & \;\;\;\; \;\;128 \;\;\;\;\;\;& \;\;\;\; \;\;256 \;\;\;\;\;\; &\;\;\;\; \;\;512 \;\;\;\;\;\; \\ \cmidrule{2-6}
                 Accuracy &  0.826& 0.835 & 0.859& 0.878 &0.897\\
                 F1-score &  0.656& 0.672 & 0.734 & 0.769 &0.816\\
            \bottomrule
        \end{tabular}
        \subcaption{Dimensionality.}
        \label{tab:compare_e}
    \end{subtable}
\end{table}

\subsection{Comparison with Existing State-of-the-Art Methods}
In this subsection, we compare the performance of the proposed FG-SSL with existing methods on the ISIC2018, APTOS2019, and ISIC2017 datasets. Table~\ref{entire_experimental} shows the experimental results from all three datasets. In previous studies, contrastive learning\cite{yang2022proco} using negative sampling achieved the best performance on the APTOS2019 and ISIC2018 datasets. Negative sampling strategy\cite{he2020momentum} that relies on large batch sizes are unsuitable for medical datasets due to data scarcity. As a result, advances in performance utilizing such methods are frequently limited in medical tasks. Yet, because our FG-SSL method does not require a high batch size, it performs well compared to state-of-the-art methods in the small sample size of the medical image recognition task, as shown in Table ~\ref{entire_experimental}.

\subsection{Ablation Study} 
We conduct three ablation studies to demonstrate the efficacy of the proposed FG-SSL.
Our models are trained in the same environment as in Sec.~\ref{details} and evaluated using the ISIC2018 Dataset in this ablation study.

\begin{figure*}[t]
\centering
\includegraphics[width=0.995\textwidth]{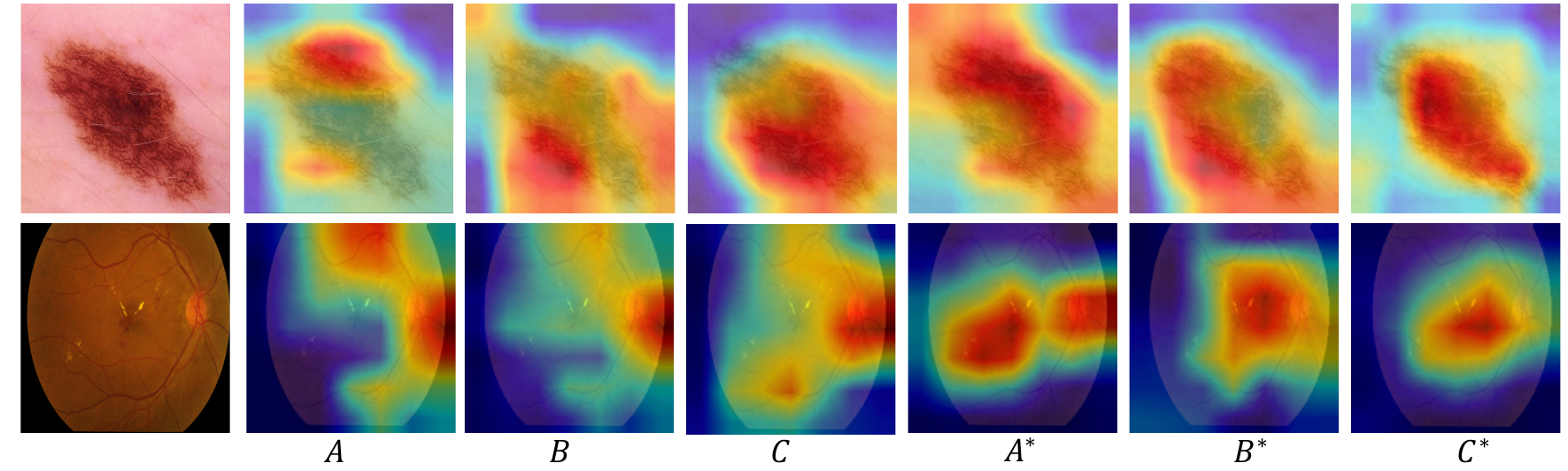}
\caption{
Grad-CAM\cite{selvaraju2017grad} visualization. A*, B*, and C* represent the result of sequentially adding the Rotation\cite{feng2019self}, SimSiam\cite{chen2021exploring}, and BarlowTwins\cite{zbontar2021barlow} methods with the proposed fine-grained self-supervised learning. At the same time, A, B, and C are taken from the baseline networks without our self-supervised learning.
}
\label{fig:gradcam}
\end{figure*}
            
\begin{figure*}[t]
\centering
\includegraphics[width=0.985\textwidth]{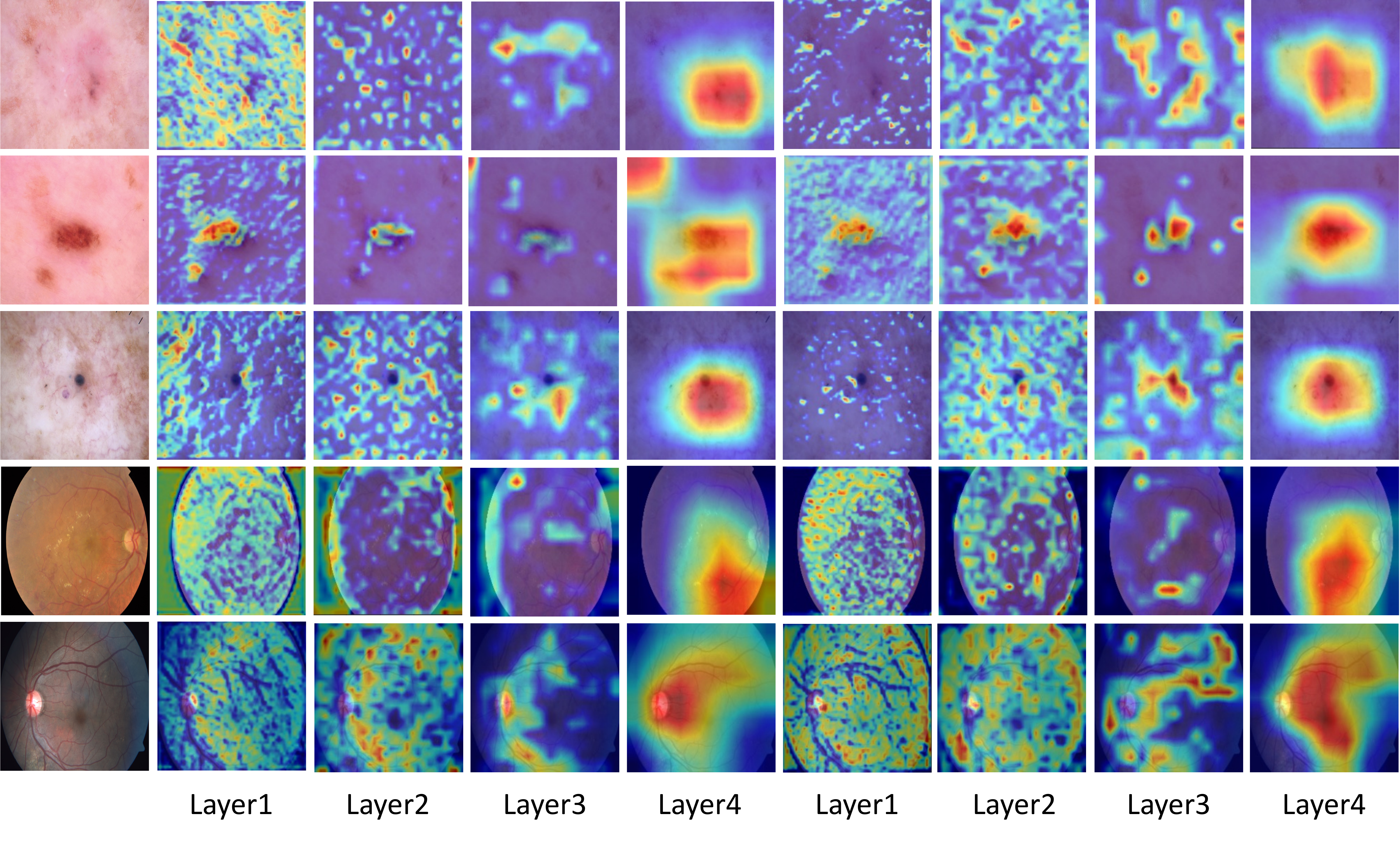}
\caption{Layer-wise GradCAM visualization. The four left columns of Layer1, 2, 3, and 4 are from a baseline network, whereas the four right columns are from our FG-SSL method.
}
\label{fig:temp}
\end{figure*}

\textbf{Quantitative analysis}: We visualize the Grad-CAM\cite{selvaraju2017grad} to evaluate the proposed method as shown in Fig.~\ref{fig:gradcam} qualitatively.   
In this visualization, we confirm that our progressive learning (A*, B, and C*) more focus on the subtle lesion region, and further, the BarlowTwins C* highlights it best. 
Further, we visualize the GradCam for each layer and found that FG-SSL more focus on subtle regions than the baseline method as show in Fig.~\ref{fig:temp}.

\textbf{Factor analysis}: 
To show the effectiveness of the progressive Jigsaw puzzles, we compare ours with other self-supervised learning methods in \Cref{tab:compare_a}. We confirm that our FG-SSL performs well compared to SimSiam and Rotation methods, while our multi-level setting also improves the performance.
For evaluating our progressive learning strategy, we compare different settings as shown in \Cref{tab:compare_b}. As the optimized hierarchical grid setting of PMG~\cite{du2020fine}, we achieve the best result on $(8, 4, 2, 1)$ among various settings. 
We also demonstrate that the proposed FG-SSL is insensitive to batch size, unlike the contrastive learning algorithm with the negative sampling (\ie MoCoV2~\cite{chen2020improved}), as shown in \Cref{tab:compare_c}. 
Additionally, we validate the performance of our FG-SSL regarding different channel dimensionality as shown in \Cref{tab:compare_e}. 
This result is significant in that our method will work well for a small sample size of medical datasets.
To evaluate the effectiveness of the component of our self-supervised learning method, we carry out an ablation study in \Cref{tab:compare_d}. When using Jigsaw puzzle and PMG, it shows competitive performance compared to ProCo\cite{yang2022proco}. 
In addition, we validate that the proposed FG-SSL method combining Jigsaw puzzle, PMG, and BarlowTwin improves the performance significantly.

\section{Conclusion}
This paper introduces FG-SSL, a novel method for classifying subtle lesions from medical images. Our FG-SSL learns fine-grained self-supervision from progressive Jigsaw puzzles and fine-tunes it with the PMG method.
Using the ISIC2018, APTOS2019, and ISIC2017 datasets, it is demonstrated that the proposed FG-SSL method performs well compared to the existing state-of-the-art methods. The efficiency of the proposed method is shown in the ablation study through extensive experiments. We validate the Jigsaw puzzle method on medical images in learning fine-grained features even without labels. We believe that our FG-SSL method will be useful in the future medical imaging community.
\bibliography{mybib}
\end{document}